\documentclass{article}
\usepackage{spconf,amsmath,epsfig}
\usepackage{amsmath}
\usepackage{multirow}
\usepackage{array}
\usepackage{booktabs}
\usepackage{url}
\usepackage[ruled,vlined]{algorithm2e}
\usepackage{balance}
\usepackage{verbatim}

\begin{document}\sloppy

\def\x{{\mathbf x}}
\def\L{{\cal L}}

\title{CODA: Counting Objects via Scale-aware Adversarial Density Adaption}
%
\name{Li Wang$^1$, Yongbo Li$^2$, Xiangyang Xue$^1$}
\address{$^1$Fudan University, $^2$Megvii Inc (Face++)\\
\{wangli16, xyxue\}@fudan.edu.cn, liyongbo@megvii.com}

\maketitle

\begin{abstract}
Recent advances in crowd counting have achieved promising results with increasingly complex convolutional neural network designs. However, due to the unpredictable domain shift, generalizing trained model to unseen scenarios is often suboptimal. Inspired by the observation that density maps of different scenarios share similar local structures, we propose a novel adversarial learning approach in this paper, i.e., CODA (\emph{Counting Objects via scale-aware adversarial Density Adaption}). To deal with different object scales and density distributions, we perform adversarial training with pyramid patches of multi-scales from both source- and target-domain. Along with a ranking constraint across levels of the pyramid input, consistent object counts can be produced for different scales. Extensive experiments demonstrate that our network produces much better results on unseen datasets compared with existing counting adaption models. Notably, the performance of our CODA is comparable with the state-of-the-art fully-supervised models that are trained on the target dataset. Further analysis indicates that our density adaption framework can effortlessly extend to scenarios with different objects. \emph{The code is available at https://github.com/Willy0919/CODA.}

\end{abstract}
\section{Introduction}
\begin{figure}[t]
  \centering
  \includegraphics[width=0.9\linewidth]{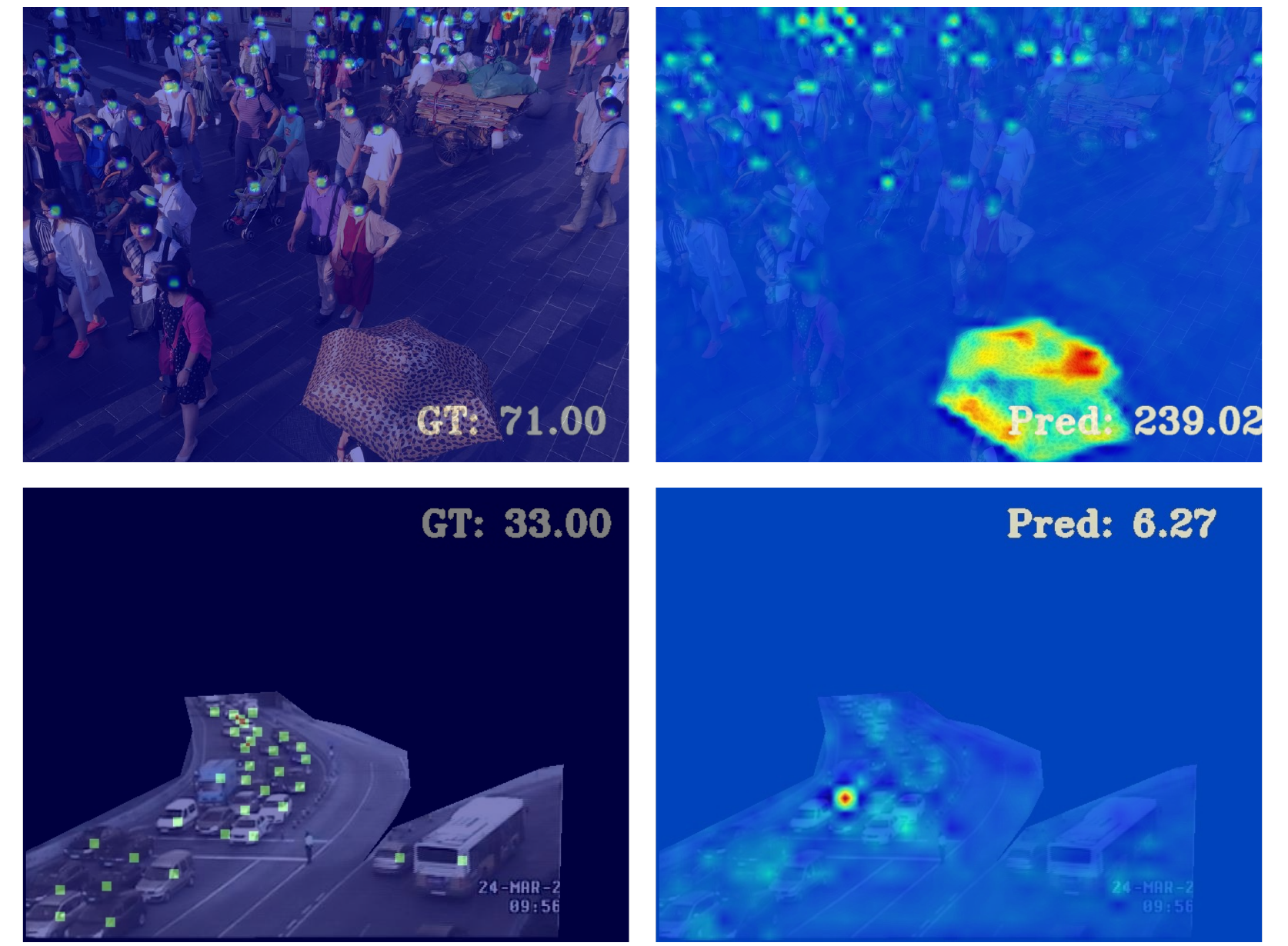}
  \caption{Two failed generalization examples of a trained counting model to unseen scenarios. The utilized counting model is trained with annotations on crowd dataset ShanghaiTech \cite{Zhang:2016fr} Part A. The unseen images are from datasets of ShanghaiTech Part B (top) and Trancos \cite{guerrero2015extremely} (bottom), respectively. GT: ground truth. Pred: estimated count. }
  \label{fig:introductiongt}
  \vspace{-0.1in}
\end{figure}


With the growing deployment of surveillance video cameras, object counting, or estimating the object counts from video frames, has been a critical functionality for various traffic and public security applications. Promising performances have been achieved under full-supervised manner by designing increasingly complex network structures under labor-intensive datasets \cite{li2018csrnet,shen2018crowd,wang2018crowd}. However, the performances degrade heavily when trained models are employed to some unseen scenarios, which limits the massive real-world applications. The challenges in the generalizing process mainly come from the extreme volatility of counts, background interferences and appearances of objects, etc. Figure \ref{fig:introductiongt} gives two failed examples of generalizing trained model to unseen images directly.  


To address the issues above, several semi- and un- supervised approaches have been proposed for crowd counting adaption from source domain to target domain. Methods of FA \cite{Loy:2013dt} and GPTL \cite{Liu:2015ie} utilize a few extra labelled samples from target datasets to optimize the current regression counting algorithms. Zhang et.al. \cite{Zhang:2015id} propose to retrieve training images similar with test scenes and adapt their pre-trained CNN model with these retrieved images. However, these existing adaption methods usually focus on the counting value regression process and ignore the spatial density information, which limits the estimation performance of trained models. Besides, additional labelled samples with representative information are also needed for unseen images. For unpredictable scenarios, how to achieve object counting with an unsupervised manner is still an open problem.  

\begin{figure*}[t]
  \centering
  \includegraphics[width=.9\linewidth]{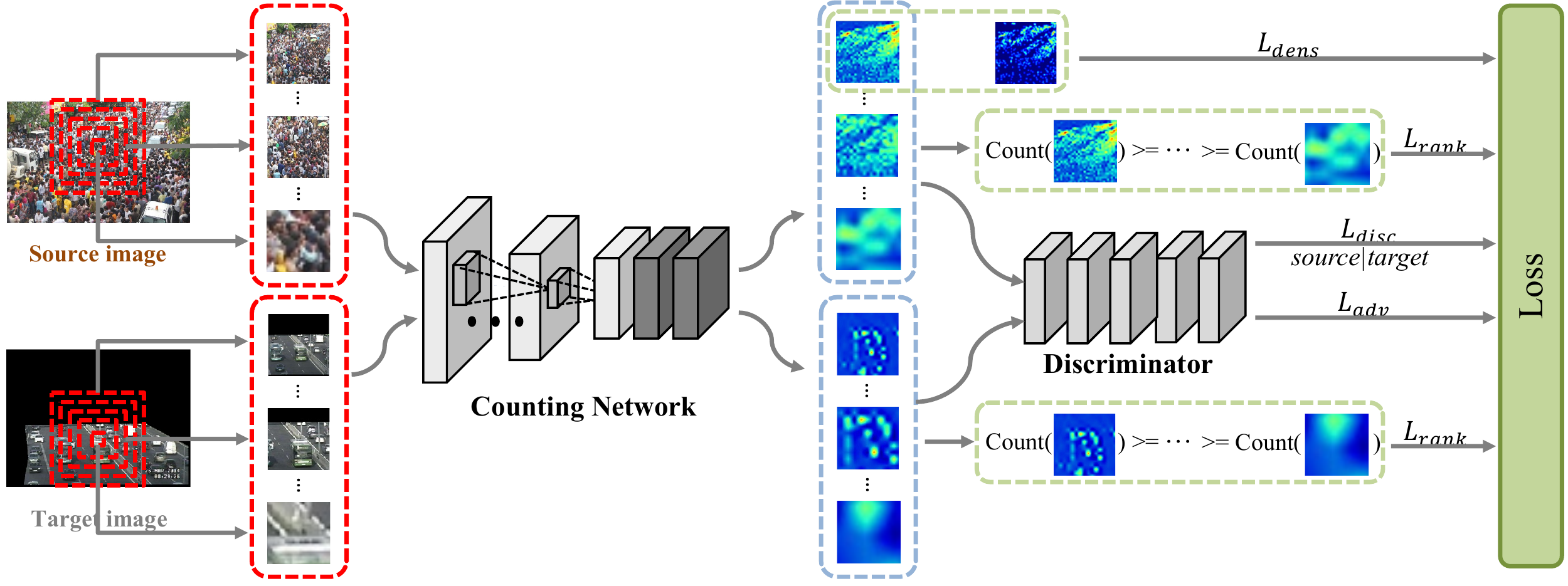}
  \caption{An overview of our CODA at density adaption stage. $Count(*)$ produces the integration of current density map. Counting Network is firstly pre-trained on source datasets only using $L_{dens}$ , and then reused as a generator at density adaption stage. At this stage, $L_{dens}, L_{disc}, L_{adv}$ and $L_{rank}$ are calculated to adapt the target predicted density maps to be more precise.}
    \vspace{-0.1in}
  \label{fig:architecture}
\end{figure*}

Although counting objects vary dramatically in appearance, the desired density maps share many local similarities such as high values in crowd area and the ignorance of background. Based on this observation, we formulate a novel adversarial framework named CODA for unsupervised density adaption across different scenarios. Unlike previous works, we propose to combine the spatial density information with the overall count value constraint, which further improves counting performance. Specifically, our CODA consists of two stages. At the first stage, a basic counting network (CN) is pre-trained with the source dataset to generate high quality density maps. At the second stage, we adapt the pre-trained model on the target datasets, i.e., unseen images without annotations, and the adversarial learning is applied. Furthermore, considering that object scale and density vary in different scenes, we take pyramid patches with multi-scales as input. And a ranking constraint across different scales is introduced to our framework to produce consistent object counts.
Experiments on common used datasets are conducted to verify the efficiency of our method. Although unsupervised, our method achieves a better performance than existing semi-supervised counting adaption models. Notably, we also demonstrate competitive adaption results compared with state-of-the-art fully-supervised models that are trained on the target dataset. The experiments also indicate the effortless density adaption across different object categories datasets.

\section{Related works}
A number of studies on object counting have been demonstrated to solve real-world problems. Traditionally, detection-based counting methods are straightforward, which utilize off-the-shelf detectors \cite{li2008estimating} to detect and count target objects in images or videos. However, the highly occluded and extremely small objects limit the performance of detectors. Therefore, regression-based approaches such as Weighted VLAD \cite{sheng2016crowd} and MoCNN \cite{kumagai2017mixture} are proposed to overcome the occlusion problem by constructing a map connecting image features and head counts. With excellent feature extraction and count value regression algorithms, these approaches achieve superior performances in several datasets. Recently, density-based counting methods \cite{lempitsky2010learning,Zhang:2016fr,li2018csrnet,shen2018crowd} have outperformed regression-based ones and been the mainstream of crowd counting approaches. Lemptisky et al. \cite{lempitsky2010learning} firstly use continuous density maps instead of discrete count numbers as optimization objective and build a mapping between image feature and density map. MCNN \cite{Zhang:2016fr} and CSRNet \cite{li2018csrnet} propose simple but robust architectures to generated high quality density maps. RSAR module \cite{liu2018crowd} is introduced to handle the effect of both scale and rotation variation for crowd counting.
On the other hand, ACSCP \cite{shen2018crowd} applies adversarial loss to narrow the distance between ground truth and predicted density map. Liu et al. \cite{liu2018leveraging} utilize ranking information from unlabeled datasets to further assist training process.

So far, Convolutional Neural Networks (CNNs) have shown their rich representative power in training and testing on a single scenario, but generalizing trained model to unseen scenarios reveals the poor performance due to the dramatically domain shift. Therefore, several semi-supervised counting algorithms have been proposed to solve the issues in domain shift and time-consuming labor annotations during counting adaption. FA \cite{Loy:2013dt} generalizes one scenario to another scenario with a small number of labelled samples by exploiting the underlying manifold structure of target crowd data. Based on Bayesian model adaptation of Gaussian processes, GPTL \cite{Liu:2015ie} uses limited annotated target domain samples to adapt Gaussian kernel. Besides, Crowd CNN \cite{Zhang:2015id} performs candidate scene and local patch retrieval on training data in the light of test data to further fine-tune the pre-trained model. 
\section{Approach}
In this section, we introduce our proposed density adaption approach, the overall pipeline is demonstrated in Figure \ref{fig:architecture}.
Notably, in the following descriptions, $I^t$ and $I^s$ represent image patches from target dataset and source dataset, $I^{t*}$ and $I^{s*}$ denote a pyramid of patches (containing the original one) cropped on origin patch $I^t$ and $I^s$ respectively.

\subsection{Counting Model Generation}

\vspace{0.05in}\noindent \textbf{Counting Network (CN)} Counting Network is the basic network to generate the density map for input patches. The network is shown in the top of Figure \ref{fig:architecture}. Inspired by CSRNet \cite{li2018csrnet}, we use VGG-16 \cite{simonyan2014very} structure as the basic feature extraction network. The final pooling layer and the following two fully connected layers in original VGG-16 are replaced by two dilated convolutional layers \cite{yu2015multi}. Moreover, a fully convolutional layer after two dilation layers is constructed to generate final density map. Dilation parameters for last two layers are set to $4$, and kernel sizes in all layers are set to $3\times3$ to reduce complexity.

Our model chooses the common Euclidean distance to measure the distance between ground truth and predicted density map. The density loss function for basic CN is described as the following:
\begin{equation}
  L_{dens}(I^s) = \frac{1}{2N} \sum_{i=1}^{N} || \hat C(I_i^s;\theta_{CN}) - C(I_i^s) ||^2\label{countloss}, \\
\end{equation}

\noindent where $\theta_{CN}$ is the counting network parameters, $N$ is the number of training batch size and $I^s_i$ is the $i$-th patch from source dataset. $C(*)$ and $\hat C(*)$ produce ground truth density map and network predicted density map respectively.

\subsection{Density Adaption}

To perform density adaption process, we reuse the CN as density map generator, and another discriminator is added to carry out adversarial learning during adaption process.

\vspace{0.05in}\noindent \textbf{Discriminator} After training CN on source dataset, we adopt a pre-trained counting model for the following adversarial learning. A Discriminator is introduced to take responsibility for distinguishing between the density maps generated by source image patch and target image. Here we use the same Discriminator proposed by Tsai et al. \cite{tsai2018learning}. It contains five convolutional layers with stride of 2 and kernel size $4\times4$, the channels of each layer are \{64, 128, 256, 512, 1\} respectively.

In this stage, Discriminator accepts a pyramid of patches as inputs to train, and cross-entropy loss used to measure classification error for each patch is as:
\begin{equation}
\begin{aligned}
	L_{disc}(I^{s*},I^{t*}) = &- \frac{1}{2NS} \sum_{i=1}^{N}\sum_{j=1}^{S}[z\log(\hat D(\hat C(I_{ij}^{s*}|I_{ij}^{t*});\theta_{D})) \\ 
	& + (1 - z)\log(1 - \hat D(\hat C(I_{ij}^{s*}|I_{ij}^{t*});\theta_{D}))]\label{discloss}
\end{aligned}
\end{equation}
\noindent where $\hat D(*)$ produces the class label (0 or 1) for current sample, $\theta_{D}$ is the parameters of Discriminator, and $S$ is the number of scaled patches. $z=0$ indicates that the current input sample is generated by target patch while $z=1$ indicates the sample is generated by source patch.

\vspace{0.05in}\noindent \textbf{Scale-aware Adversarial Learning} To make target predicted density maps more similar to source ones, we append an adversarial loss to make Discriminator mistake target density maps for the source ones. Since even scenarios with very different density distributions can often share many similar local density patterns, local image patches can be critical for density adaption. To take advantage of this, a scale-aware framework is formulated to adapt object scales and density distributions, and the overall count values consistency is imposed on adversarial learning based on this strategy. As shown in the left of Figure \ref{fig:architecture}, a pyramid of patches from source or target dataset is taken as inputs of CN which shares the same center across different scales. Given these source and target patches, CN resizes them into the same size and forwards each patch to obtain corresponding output density maps. An adversarial and a ranking loss are enforced to optimize the overall network. For adversarial loss, we use cross-entropy loss to confuse the Discriminator, as described:
\begin{equation}
	 L_{adv}(I^{t*}) = - \frac{1}{2NS} \sum_{i=1}^{N}\sum_{j=1}^{S}\log(\hat D(\hat C(I_{ij}^{t*};\theta_{CN});\theta_{D})) \label{advloss}
\end{equation}

\noindent where we only compute the adversarial loss for target patches.

To restrict count value consistency, we leverage the intrinsic relationships between these scaled input patches. Obviously, each patch from both datasets satisfies the constraint that count value in bigger patch is equal or greater than that in any included smaller one. Therefore, we use ranking loss to implement the constraint, which is computed as:
\begin{equation}
	 L_{rank}(I^{s*},I^{t*}) = \sum_{i>j}^S\sum_{j=1}^{i}max(0, \hat n_j- \hat n_i + \varepsilon)\label{rankloss}
\end{equation}
\begin{equation}
	 \hat n_i = \int \hat C(I_i^{s*}|I_i^{t*};\theta_{CN})\label{count}
\end{equation}

\noindent where $\hat n_*$ is the integration of predicted density map for the $*$-th patch. $\varepsilon$ is the real difference between two patches, the count values for two patches can be equal in our case so we set $\varepsilon = 0$. Note that the larger index in $\hat n_*$ represents the output of bigger patch. 

\vspace{0.05in}\noindent \textbf{Training process} The final loss function for our model is
\begin{equation}
\begin{aligned}
	L(I^s,I^t,I^{s*},I^{t*}) = &L_{dens}(I^s) + \lambda_1 \sum_{j=1}^{S}L_{disc}(I^{s*},I^{t*}) +\\
	 &\lambda_2 L_{adv}(I^{t*}) + \lambda_3 L_{rank}(I^{s*},I^{t*})\label{finalloss}
\end{aligned}
\end{equation}

\noindent where $\lambda_1$, $\lambda_2$ and $\lambda_3$ balance different losses and are all set to 0.001. Based on Eq. \ref{finalloss}, the training target is to optimize:
\begin{equation}
	 \max \limits_{D} \min \limits_{G} L(I_s,I_t,I^{s*},I^{t*})
\end{equation}

\noindent where $\mathit{G}$ represents the CN, and $\mathit{D}$ is the Discriminator. The equation represents that network tends to minimize the density loss for source patches, while maximize the differences for Discriminator to recognize the predicted density map produced by source patches or target patches.

To obtain the final adaption model, we firstly pre-train CN on the source dataset using $L_{dens}$. Then the overall scale-aware adversarial learning process is trained end-to-end. More specially, we feed a pyramid of patches $I^{s*}$ and $I^{t*}$ into the pre-trained model, and $L_{I^s,I^t,I^{s*},I^{t*}}$ is calculated to further optimize CN and Discriminator.

\vspace{0.05in}\noindent \textbf{Implementation details.} All network input patches are resized as $512\times512$ with 3 channels. For data augmentation and ground truth density map generation, we apply the same strategy as in MCNN \cite{Zhang:2016fr}. A scale-aware pyramid of patches is cropped on initial patch's center with the same aspect ratio, and the scales are set as $\{0.8,0.6,0.4\}$ times of the initial patch to make a balance between time and accuracy. The CN is trained using the Stochastic Gradient Descent (SGD) optimizer with learning rate as $10^{-6}$. The Discriminator uses Adam optimizer \cite{kingma2014adam} with initial learning rate as $10^{-3}$ and decreases using polynomial decay with power of 0.9, weight decay is set to $10^{-4}$ while momentum is set to 0.9 and 0.99.

\section{Experiments}
In this part, we set three experiments to verify the effectiveness of CODA: UCSD2Mall, ShanghaiTech A2B and A2Trancos ('DatasetA2DatasetB' means the experiment trained on DatasetA and adapted to DatasetB).
The metrics we use include Mean Absolute Error, $\mathit{MAE}$=$\frac{1}{N} \sum_{i=1}^{N} | c_i - \widehat c_i |$; Mean Squared Error, $\mathit{MSE}$=$\sqrt{\frac{1}{N}  \sum_{i=1}^{N} ( c_i - {\widehat c}_i )^2}$, and additionally $\mathit{{GMAE}(L)}$ = $\frac{1}{N} \sum_{i=1}^{N} (\sum_{l=1}^{4^L}|c_{i}^{l} - {\widehat c}_{i}^{l}|)$. For metrics, $N$ is the number of the testing images, $c_i$ (or $c_{i}^{l}$) and $\widehat c_i$ (or ${\widehat c}_{i}^{l}$) are the ground truth and the predicted count number for the $i$-th test image (in region $l$). For the special level $L$, $\mathit{GMAE(L)}$ divides image into a grid of $4^L$ non-overlapping subregions, and the error is calculated as the sum of $\mathit{MAE}$ in each subregion. Note that $\mathit{GMAE(0)}$ is equivalent to $\mathit{MAE}$ metric. Therefore, the lower $\mathit{MAE/MSE/GMAE}$ is calculated, the better performance is achieved. Like other works, we evaluate $\mathit{MAE/MSE}$ on UCSD2Mall and A2B, while $\mathit{GMAE}$ is calculated on A2Trancos.

\subsection{Comparing with State-of-The-Art}
\vspace{0.05in}\noindent \textbf{UCSD2Mall.} In consideration that, existing counting adaption works use UCSD dataset \cite{chan2008privacy} as source dataset and Mall dataset \cite{chen2012feature} as target dataset, we also demonstrate the results assembled with the same setting.

UCSD dataset \cite{chan2008privacy} collects 2000 frames of a video from a surveillance camera recording the pedestrians along walkway, and the frames are chosen at 10 fps from the video and down sampled as a size of $158\times238$. Meanwhile, Mall Dataset consists of 2000 frames with a fixed resolution of $320\times240$.
We verify our model comparing with existing counting adaption methods, including FA \cite{Loy:2013dt}, HGP \cite{yu2005learning} and GPTL \cite{Liu:2015ie}. We follow the settings in \cite{chan2008privacy}, and demonstrate the results on testing frames 801-2000 on Mall dataset. 

\begin{table}[h]
\small
\begin{center}
\begin{tabular}{c|ccc }
\hline
\multirow{2}{*}{Methods} &\multicolumn{3}{c}{UCSD2Mall}\\
\cline{2-4}
&Target supervision&MAE&MSE\\
  \hline
  \hline
  FA   &semi&	7.47   &  -  \\
  HGP  &semi&	4.36  & -   \\

  GPTL  &semi&	3.55   &  - \\
  \hline
  \hline
  Baseline   &no&4.00   &  5.01 \\
  CODA  &no&  \textbf{3.38}   &  \textbf{4.15}   \\
  \hline
\end{tabular}
\end{center}
\vspace{-0.1in}
\caption{$\mathit{MAE}$/$\mathit{MSE}$ comparison results under UCSD2Mall setting. Note that Target supervision represents if method is directly trained on target dataset with annotations.}
\label{tab:ucsd2mall}
\vspace{-0.1in}
\end{table}

As illustrated in Table \ref{tab:ucsd2mall}, our unsupervised CODA outperforms all semi-supervised adaption methods, improving $\mathit{MAE}$ performance from 3.55 to 3.38. This is a strong evidence that our adversarial learning adapts a better series of parameters than previous semi-supervised counting adaption methods. Baseline is the testing results without density adaption process on target dataset. Obviously, our CODA improves $\mathit{MAE/MSE}$ by 15.5\%/17.2\% after adaption training.

\vspace{0.05in}\noindent \textbf{ShanghaiTech Part A2B.} We also compare our CODA with the current state-of-the-art full-supervised approaches on ShanghaiTech dataset \cite{Zhang:2016fr} which is one of the largest datasets available in terms of annotation. It contains 1,198 annotated images with a total of 330,165 people. The dataset consists of two subsets: Part A and Part B. Part A has 482 images collected from the Internet while Part B includes 716 images captured from downtown Shanghai. Our CODA trains CN on Part A and conducts density adaption on Part B without annotations while full-supervised approaches train and test only on Part B. We follow the settings in \cite{Zhang:2016fr} using the 316 out of 716 images as testing set to fairly compute the $\mathit{MAE}$ and $\mathit{MSE}$.
 
 \begin{table}[h]
\small
\begin{center}
\begin{tabular}{c|ccc }
\hline
\multirow{2}{*}{Methods} &\multicolumn{3}{c}{A2B}\\
\cline{2-4}
&Target supervision &MAE&MSE\\
  \hline
  \hline
  MCNN  &yes&	26.4   &  41.3  \\

  ACSCP   &yes&	17.2   &  27.4 \\
  CSRNet   & yes&\textbf{10.6}   &   \textbf{16} \\
  \hline
  \hline
  Baseline   &no&27.3   &  36.2 \\
    CODA   &no&\textbf{15.9}   &  \textbf{26.9}  \\
  \hline
\end{tabular}
\end{center}
\vspace{-0.1in}
\caption{$\mathit{MAE}$/$\mathit{MSE}$ comparison results under ShanghaiTech Part A2B setting.}
\label{tab:partb}
\vspace{-0.1in}
\end{table}
In Table \ref{tab:partb}, we report performances of our model and the other state-of-the-art full-supervised approaches, including MCNN \cite{Zhang:2016fr}, ACSCP \cite{shen2018crowd} and CSRNet \cite{li2018csrnet}. Interestingly, our CODA delivers competitive performance without any annotations on target dataset compared with other full-supervised methods, which achieves lower $\mathit{MAE}$ and $\mathit{MSE}$ results than all methods except CSRNet. It also confirms the effectiveness of our density adaption that CODA improves Baseline performance from 27.3/36.2 to 15.9/26.9.

\vspace{0.05in}\noindent \textbf{ShanghaiTech Part A2Trancos.} 
In this section, we also present the adaption results across different categories. Trancos dataset \cite{guerrero2015extremely} contains 1244 images of different traffic scenes collected from real video surveillance cameras. It annotates 46796 vehicles totally and provides a region of interest for each image to specify the scope of the evaluation. We follow the pre-defined settings to evaluate on test set (421 images).

\begin{table}[h]
\small
\begin{center}
\begin{tabular}{c|p{0.36in}<{\centering}p{0.36in}<{\centering}p{0.36in}<{\centering}p{0.36in}<{\centering} }
\hline
\multirow{2}{*}{Methods} &\multicolumn{4}{c}{A2Trancos}\\
\cline{2-5}
&GMAE0&GMAE1&GMAE2&GMAE3\\
  \hline
  \hline
  Lempitsky et al.*&	13.76&16.72	&20.72&24.36\\
  Hydra-CNN*  &	10.99&13.75&16.69&19.32  \\
  CSRNet*  &	\textbf{3.56}&\textbf{5.49}&\textbf{8.57}&\textbf{15.04} \\
  \hline
  \hline
  Baseline   &13.71&13.81&\textbf{14.52}&\textbf{15.75} \\
  CODA  & \textbf{4.91}&\textbf{9.89}&14.88&17.55	 \\
  \hline
\end{tabular}
\end{center}
\vspace{-0.1in}
\caption{$\mathit{GMAE}$ comparison results under ShanghaiTech Part A2Trancos setting. * denotes that models are trained and tested on target dataset with full-supervision. }
\label{tab:trancos}
\vspace{-0.1in}
\end{table}

Table \ref{tab:trancos} presents the results compared with the state-of-the-art approaches:  Lempitsky et al. \cite{lempitsky2010learning},  Hydra-CNN \cite{onoro2016towards} and CSRNet \cite{li2018csrnet}. As shown in table, Baseline results are abnormal which show little difference between $\mathit{GMAEs}$, so we visualized the estimated density maps and found that the count value for each image is from 0 to 5, which are completely failed on predictions. By contrast, after our density adaption process, our CODA enhances performance up to 64.2\% in $\mathit{GMAE0}$ compared with Baseline and achieves a great performance close to the best. It is concluded that our CODA has a strong capability of density adaption which not only preforms well when adapting crowds, but also presents great effectiveness on adaption across categories.

\begin{figure*}[t]
  \centering
  \includegraphics[width=.99\linewidth]{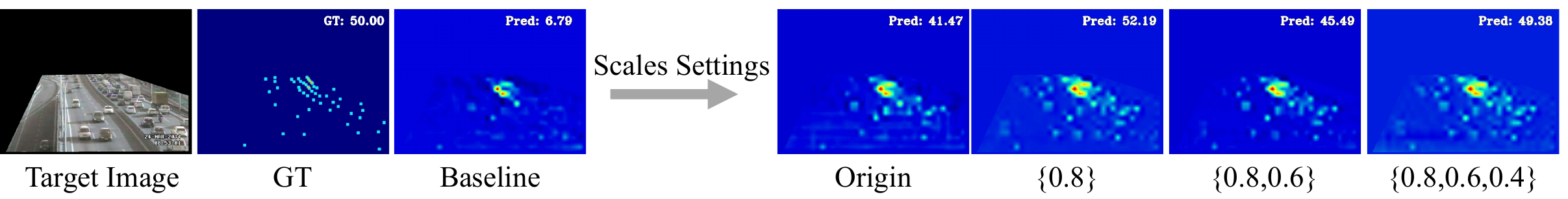}
  \caption{Adaption samples for different input patch scales under A2Trancos setting.}
  \vspace{-0.1in}
  \label{fig:ablation}
\end{figure*}
\begin{figure*}[t]
  \centering
  \includegraphics[width=.99\linewidth]{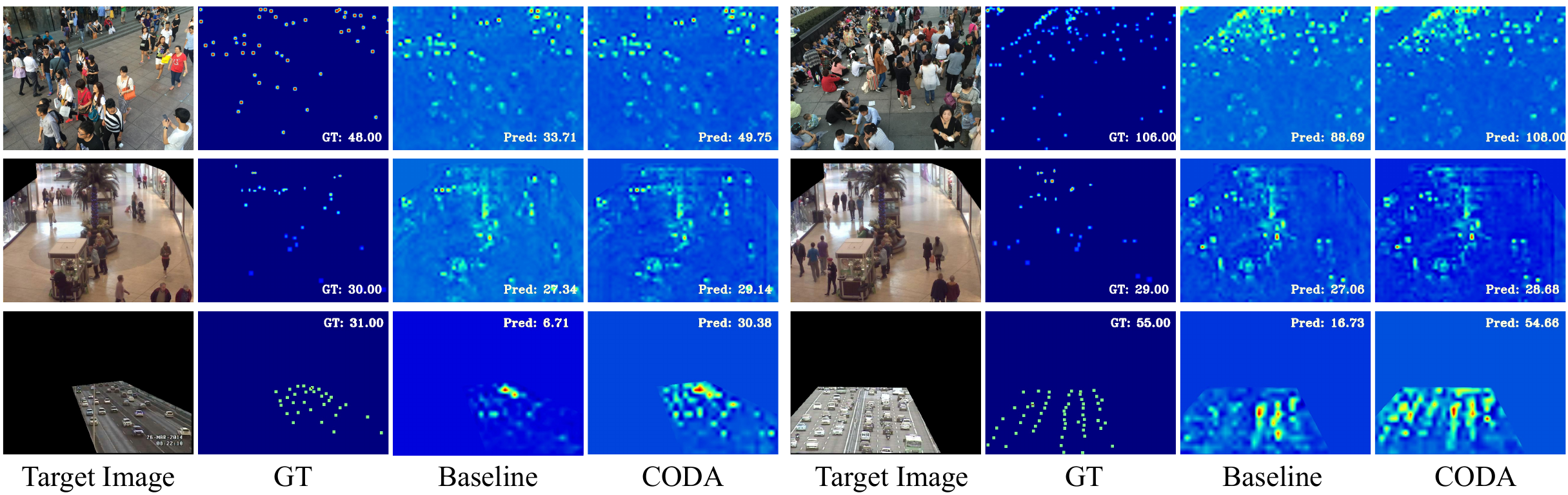}
  \caption{Qualitative adaption results. Results from top to down are under UCSD2Mall, A2B and A2Trancos settings respectively.}
  \vspace{-0.1in}
  \label{fig:test}
\end{figure*}

\subsection{Discussion}
In this part, we explore the influence of scale size settings for scale-aware adaption strategy. 

\begin{table}[h]
\small
\begin{center}
\begin{tabular}{c|p{0.36in}<{\centering}p{0.36in}<{\centering}p{0.36in}<{\centering}p{0.36in}<{\centering}}
\hline
\multirow{2}{*}{Scale settings} &\multicolumn{4}{c}{A2Trancos}\\
\cline{2-5}
 &GMAE0&GMAE1&GMAE2&GMAE3\\
  \hline
  \hline
  origin  &	6.89&12.81&15.82&19.07  \\
  \{0.8\}& 5.54&\textbf{8.35}&16.30&18.47\\
    \{0.8,0.6\}& 5.31&9.66&\textbf{12.08}&18.55\\
    \{0.8,0.6,0.4\}& \textbf{4.91}&9.89&14.88&\textbf{17.55}\\
  \hline
\end{tabular}
\end{center}
\vspace{-0.1in}
\caption{$\mathit{GMAE}$ results on density adaption from Part A to Trancos dataset under different scale settings.}
\label{tab:ablationrank}
\vspace{-0.1in}
\end{table}

$\mathit{GMAE}$ results under different input patch scale settings are shown in Table \ref{tab:ablationrank}. We set five options: origin represents that we train network without scale-aware strategy and ranking loss; the others represent scaled sizes for cropped patches (including original one). It is pointed that with the number of scales increases, the density perception of network is stronger, and constraint becomes stricter. When setting scales at \{0.8, 0.6, 0.4\}, the $\mathit{GMAE0}$ and $\mathit{GMAE3}$ perform best while $\mathit{GMAE1}$ and $\mathit{GMAE2}$ achieve a relative lower error. We also visualize qualitative testing samples in Figure \ref{fig:ablation} for these different scale settings. Obviously, the option \{0.8, 0.6, 0.4\} performs best compared with others. More qualitative adaption results can be seen in Figure \ref{fig:test}.

\section{Conclusion}
In this paper, we propose an unsupervised density adaption algorithm in counting objects, i.e., CODA. Since the counting annotation consumes a lot of manpower and the enormous domain gaps exist between different scenarios, the performance of generalizing trained model to unseen scenarios straightly is unsatisfied. Owing to the local structure similarity between different scenarios, we propose an adversarial learning algorithm to adapt trained model to unseen scenarios, and further improve quality of density maps using scale-aware training and counting consistency between multi-scale patches. Qualitative results on three experiment configures verify the effectiveness of our model on adapting across crowd scenarios, and it also indicates that our density adaption approach can effortlessly extend to scenarios with different object categories.


\section{Acknowledgement}
This work was supported in part by National Key R\&D Program of China (No.2017YFC0803700), NSFC under Grant(No.61572138 \& No.U1611461) and STCSM Project under Grant No.16JC1420400.

\balance
\bibliographystyle{IEEEbib}
\bibliography{icme2019template.bbl}

\end{document}